\documentclass[sigconf, nonacm=true]{acmart}
\usepackage{microtype}
\usepackage{graphicx}
\usepackage{subfigure}
\usepackage{booktabs} 
\usepackage{amsmath}
\usepackage{amsthm}
\usepackage{collcell}
\usepackage{tabularx}
\usepackage{tikz}
\usepackage{xcolor}
\usepackage{bbm}
\usepackage[linesnumbered,ruled,vlined]{algorithm2e}
\usepackage[labelfont=bf,format=plain,justification=centering,singlelinecheck=false]{caption}
\usepackage{ragged2e}
\usepackage{multirow}

\newcommand\inlineheader[1]{\texttt{\textbf{#1}}}

\DeclareMathOperator*{\argmin}{arg\,min}

\AtBeginDocument{%
  \providecommand\BibTeX{{%
    \normalfont B\kern-0.5em{\scshape i\kern-0.25em b}\kern-0.8em\TeX}}}

\setcopyright{none}
\copyrightyear{2021}
\acmYear{2021}
\acmDOI{}

\acmConference[]{}
\acmBooktitle{}
\acmPrice{}
\acmISBN{}

\makeatletter
\def\@copyrightspace{\relax}
\makeatother

\begin{document}

\definecolor{brewer@qual_dark_1}{RGB}{ 27,158,119}
\colorlet{up}{brewer@qual_dark_1}
\definecolor{brewer@qual_dark_4}{RGB}{231, 41,138}
\colorlet{down}{brewer@qual_dark_4}
\definecolor{brewer@qual_dark_4}{RGB}{138, 138,138}
\colorlet{equal}{brewer@qual_dark_4}

\newcommand\up[1]{\textcolor{up}{$\uparrow$#1}}
\newcommand\down[1]{\textcolor{down}{$\downarrow$#1}}
\newcommand\eq[1]{\textcolor{equal}{--#1}}

\newcommand\RR{\textit{Relation Resolution}}
\newcommand\DistMult{\textit{DistMult}}
\newcommand\RESCAL{\textit{RESCAL}}
\newcommand\ANALOGY{\textit{ANALOGY}}
\newcommand\HolE{\textit{HolE}}
\newcommand\ComplEx{\textit{ComplEx}}

\theoremstyle{definition}

\title{V-Coder: Adaptive AutoEncoder for Semantic Disclosure in Knowledge Graphs}


\author{Christian M.M. Frey}
\affiliation{%
  \institution{Ludwig-Maximilians-Universit\"{a}t,Institute for Informatics}
  \streetaddress{Oettingenstr. 67}
  \city{Munich}
  \country{Germany}}
\email{Christian.Frey@lmu.de}

\author{Matthias Schubert}
\affiliation{%
  \institution{Ludwig-Maximilians-Universit\"{a}t,Institute for Informatics}
  \streetaddress{Oettingenstr. 67}
  \city{Munich}
  \country{Germany}}
\email{schubert@dbs.ifi.lmu.de}

\renewcommand{\shortauthors}{Frey, et al.}

\begin{abstract}
Semantic Web or \emph{Knowledge Graphs} (KG) emerged to one of the most important information source for intelligent systems requiring access to structured knowledge. One of the major challenges is the extraction and processing of unambiguous information from textual data. Following the human perception, overlapping semantic linkages between two named entities become clear due to our common-sense about the context a relationship lives in which is not the case when we look at it from an automatically driven process of a machine. In this work, we are interested in the problem of \emph{Relational Resolution} within the scope of KGs, i.e, we are investigating the inherent semantic of relationships between entities within a network. 
We propose a new adapti\textbf{v}e AutoEn\textbf{coder}, called \emph{V-Coder}, to identify relations inherently connecting entities from different domains. Those relations can be considered as being ambiguous and are candidates for disentanglement. Likewise to the \emph{Adaptive Learning Theory} (ART), our model learns new patterns from the KG by increasing units in a competitive layer without discarding the previous observed patterns whilst learning the quality of each relation separately. 
The evaluation on real-world datasets of Freebase, Yago and NELL shows that the V-Coder is not only able to recover links from corrupted input data, but also shows that the semantic disclosure of relations in a KG show the tendency to improve link prediction. A semantic evaluation wraps the evaluation up. 

\end{abstract}

\ccsdesc[500]{Information systems~Clustering}
\ccsdesc[500]{Computing methodologies~Neural networks}
\ccsdesc[300]{Computing methodologies~Cluster analysis}
\ccsdesc[300]{Computing methodologies~Topic modeling}


\keywords{Semantic Web, knowledge graph, adaptive resonance theory, adaptive AutoEncoder, Relation Resolution}


\maketitle


\section{Introduction}
\begin{figure}
    \centering
        \includegraphics[width=\columnwidth]{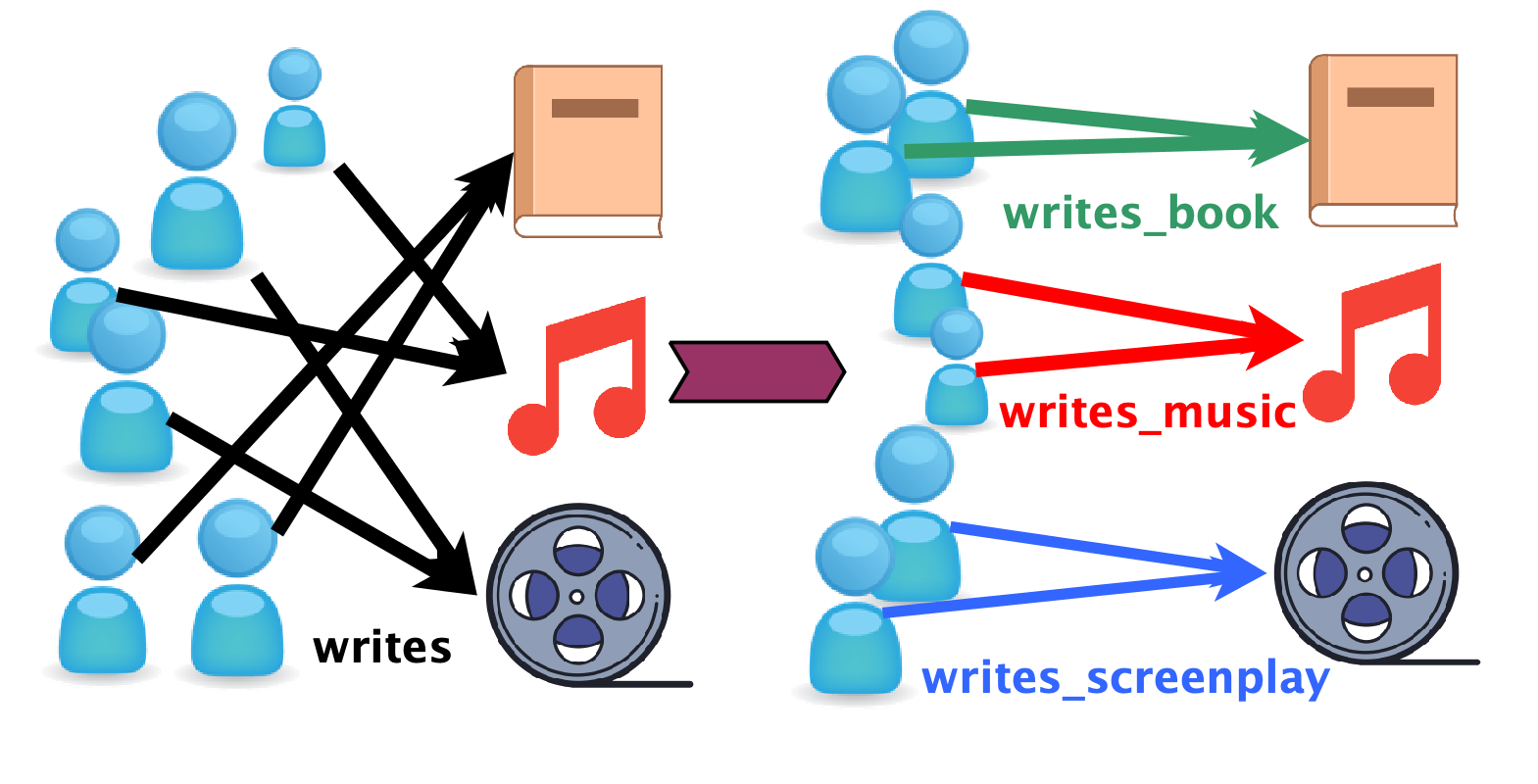}
    \caption{Semantic relation 'writes' within a KG and its disassembled semantics}
    \label{fig:motivation}
\end{figure}

In 2010 Google introduced their concept of `Knowledge Graphs' (KGs) to a broader community and has since then been used as a term in computer science articles, papers and diverse publications. Whereas graphs have always been pervasive as a major research area in Artificial Intelligence (AI), there is nowadays a need to enable machines to understand and giving them the ability to infer new knowledge. Therefore, one of the main intuition behind Knowledge Graphs was to be able to search for `things not strings' opening the gate of reasoning about entities, attributes and relationships which comes along with this kind of structured data in the form of graphs. For example, in order to enrich search queries over the web, Google developed their so-called knowledge panels in their web interface. This enables to provide additional information in a condensed form to the user where this information is extracted from the underlying knowledge graph. With Google being undoubtedly one of the pioneers in this area, other companies started using the concept of representing knowledge in forms of graph and we can observe even KG-centric startups arising for all kinds of applications such as in government, economic or non-profit and last but not least the KG ecosystem emerged to one of the major research fields in cognitive automation systems. 
An interesting problem arises from the given or inherent ontology on which the KG is built upon on. Having a hierarchical semantic tree in mind we can ask ourselves questions about the types of entities, semantic of relationships, level of granularity of resolutions for entities as well as linkages, connections across levels, level of perception we want to provide and so on. 
Whereas entity resolution is a well-studied problem \cite{Getoor-2012-ER}, we want to emphasize the resolution for relationships in this work. \textit{Relation Resolution} is the problem of devising algorithms solutions for determining linkages between two entities within a knowledge graph. Hence, we focus on the inherent semantics coming along with connections between entities being of valuable importance whenever researchers are faced with linked data, including the communities for knowledge discovery and data mining, databases and semantic web.
The most important challenge in automating \emph{Relation Resolution} is the ambiguity of the extracted information. The relationships within a knowledge graph have an inherent nature of a common-sense which allows by human perception to tell whenever two linkages are the same or not, i.e., to tell that two relationships have a different semantic between entities although the linkages are labeled with the same identifier. This phenomena can also be observed in natural language, where we can infer information just by setting it in a broader context. For example, if we just consider a relationship \textit{'write'}, we cannot infer any broader information from it with a sufficient high level of confidence. If we have the information of the context, e.g., the entities \textit{'Voltaire'} and \textit{'Candide ou l'Optimisme'}, from a human perception point of view, we automatically infer the information of the first entity (subject) being an author having written (predicate \textit{'write'}) a piece of work called \textit{'Candide ou l'Optimisme'} (object). This results from the human common-sense about nature and things we are aware of. Knowledge graphs encode information such that information can also be interpreted and extracted automatically by a machine. Therefore, the information is encoded via semantic triples having the structure \textit{(subject, predicate, object)}. Nevertheless, the predicate \textit{'write'} is not solely used to connect authors with their works, but may also connect musicians and their piece of works, directors with their screenplays or in modern terms people write blog entries and so on (c.f. figure \ref{fig:motivation}). 

Hence, from a mathematical point of view, the information provided by \textit{'write'} tries to connect several subspaces from the subjects with subspaces from the objects, namely the ones in which the various entities (authors, composers, directors, ...) lives in. The crucial part here is that these subspaces might be independent from each other resulting in the worst case in lower performance of link prediction tasks in a knowledge graph.  
Moreover, the mentioned context is given by entities in the neighborhood of a node which is connected via links carrying a specific semantic information. By that, a machine can use this semantic information and represent it. But it fails in distinguishing the inherent more fine-granular semantics of an identifier (e.g., a string). From the human perception the linkage of e.g. \textit{member\_of} can be interpreted from different angles such as connections of people to sport clubs, parties, organizations, enterprises, clans and so on. Nevertheless, a machine will deal with all entities being connected by this link in the same way even though these entities represent various and probably independent groups. In the following we will deal with this problem of \textit{Relation Resolution} in more detail and show that we are able to automatically extract ambiguous relations and separate them semantically from each other. 

\textbf{Outline.} In section \ref{sec:related_works}, we give an overview of published works being relevant for our task at hand. In section \ref{sec:fundamentals}, a summary of important definitions are given being used throughout the presentation of our framework and a short recap of the \emph{Adaptive Resonance Theory} is presented. We introduce our novel architecture, the V-Coder in section \ref{sec:adaptiveAutoEncoder}. The results of the evaluation are presented in section \ref{sec:evaluation}. Finally, section \ref{sec:conclusion} concludes this paper.

\section{Related Works}
\label{sec:related_works}

Starting from the 2010s the research community could observe a rapid growth in knowledge graph (KG) construction and applications. Not only in the number of public available datasets like, Freebase \cite{Bollacker2008}, Wordnet \cite{Miller1995} or Yago \cite{Suchanek2007} but also technically finesses weighting the inherent semantic meanings which naturally comes along with the structure of a knowledge graph. 

\noindent \inlineheader{Embedding Methods}\hfill\\
In the trend of representing high-dimensional data in a lower dimensional space knowledge graphs are processed by embeddings models for which the interested reader is referred to a thorough overview given by \cite{journals/tkde/WangMWG17}. It is worth to mention that the challenge of embeddings is to find lower dimensional manifolds by taking the information given by an entity (and its neighborhood) rather than dealing with the resolution of the entities or of the relationships. A lower-dimensional representation can then further be used for tasks like link-based clustering grouping objects in relational data being similar, hence, exploiting the \emph{homophily} property. 

{\bfseries Translational methods.} One of the pioneers work was proposed by Bordes et al. in 2013 when introducing TransE \cite{bordes13transe}. In this work the learned representation of relations served as translations in vector space. Further improvements were introduced subsequently based on this model, e.g., TransH \cite{wang14transh} introduced relation-specific hyperplanes with a normal vector, TransR \cite{lin-2015-transr} introduced relation-specific spaces where entity representations are projected into the space specific to a relation with a projection matrix from the entity space to the relation space. 
In 2018, RotatE was introduced in \cite{sun18rotate} proposing a rotation-based translational method but in complex space.

{\bfseries Semantic Matching Models.} Rather than translating vectors such that they match, semantic matching models measures the plausibility of relations by matching latent semantics of entities and relations. Yang et al. introduced DistMult in \cite{yang2015distmult} where each relation is represented as a diagonal interaction matrix between the head and tail embeddings. SimplE \cite{kazemi-2019-simplE} introduced a bilinear approach where head entity serve also as tail entities and vice versa. 
In \cite{trouillon16complex}, the authors introduced ComplEx, extending DistMult by working with embeddings in the complex space $\mathbb{C}$ such that asymmetric relations have more expressiveness notion in the model. A model which subsumed all of the mentioned semantic matching models is ANALOGY \cite{liu2017analogy}. 

{\bfseries Matching with Neural Networks.} 
A Neural Tensor Network \cite{socher13ntn} uses projection of entities in the embedding space and combines them with a relation-specific tensor. Neural Association Model works with concatenations of vector embeddings and feds them into a deep neural network such that the output of the last hidden layer and the embedding of the tail entity tells us the matching score. More recently, \cite{dettmers18conve} introduced ConVE, a convolution neural networks, graph convolutional networks were introduced by \cite{schlichtkrull18rgcn}, or deep memory networks in \cite{wang18memorynetwork}.

Nevertheless, even though the above methods represents the relational information in a lower-dimensional space, they are not dealing with the relational resolution in order to uncover the inherent semantics of a relation.

\noindent\inlineheader{Disambiguation Tasks}\hfill\\
In the community of Natural Language Processing (NLP) we can find more specific tasks taking care of the Named Entity Disambiguation (NED) or Named Entity Linking whose primary goal is to assign unique identifiers to entities being mentioned in a text. For example, Einstein won the Noble Prize in Physics in 1921. The entity mention of "Einstein" should be linked to the entity of Albert Einstein. NED can be considered as a follow-up module in a pipeline with Named Entity Recognition (NER) in first place.  For example in \cite{Zheng2012EntityDisambiguation} the authors present an approach on how to solve disentangling ambiguous entities for the Freebase dataset. 
Another framing of the problem would be \textit{Entity Resolution}\cite{Singla2006, Bhattacharya2007, Whang2013} (or \textit{Schema Matching} \cite{Rahm2001}, \textit{De-duplication} \cite{Culotta2005}, \textit{Object Identification} \cite{Tejada2001}) where the task would be to enrich the inherent ontology for the various entity types. This is of specific usage when decisions are made for all object which can be referred to the same domain rather than doing them independently for each object pair. On the other hand objects referring to the same type are merged together. 

Research groups approached the topic of knowledge graphs also from an \textit{Information Extraction }(IE) point of view which aims at generating structured graph data from text populating a knowledge base. More related to what we are concerned about, the relations, is the sub-task of \textit{Relation Extraction} (RE) which assigns the most likely type of the relation given by a knowledge base. Such alignments were proposed for example with the combination of embedding model \cite{weston-etal-2013-connecting} or with the help of latent factors given by a matrix factorization \cite{riedel-etal-2013-relation}.

Our main task of \textit{Relation Resolution} (which could also be framed as \textit{Named Relation Disambiguation} (NRD), \textit{Relation De-duplication} refers to the problem of disentangling different relations linking named entities which were extracted from a text and therefore form a knowledge base. 
A more related work is given by \cite{chen-etal-2006-unsupervised-relation} presenting an unsupervised approach for relation disambiguation in a text by an elongated k-means approach performing for each increment of $k$ a Spectral clustering.
\section{Fundamentals}
\label{sec:fundamentals}

\subsection{Adaptive Resonance Theory}
\subsubsection{\bfseries Basics of ART}\hfill\\
The Adaptive Resonance theory (ART) \cite{grossberg-76} was introduced to mimic biologically processes of how a brain learns and adapts to patterns in a constantly changing environment. Its terms 'adaptive' and 'resonance' are used to underline the functionality of the system being able to learn new patterns without discarding the previous observed patterns, resp., the old information. Therefore, the resonance in the system is regulated in the architecture with feedback. 
The ART networks are known to solve the so-called \emph{Stability-plasticity} dilemma \cite{carpenter-1988}, where stability refers to their nature of memorizing the learning and plasticity refers to the fact that they adapt towards new information. Nevertheless, plasticity can lead to instabilities in the system where new knowledge leads to a loss of previously learned information. This phenomenon is also know es \emph{catastrophic forgetting} which is addressed by the ART architectures. 

In general, ART models implement a clustering algorithm where the input is presented to the networks and the model checks whether it fits into one of the already learned clusters. If such a cluster cannot be found, a new cluster is formed. Hence, it follows an unsupervised learning approach. 



\subsubsection{\bfseries Basic Architecture of ART models}\hfill\\
The essence of the ART is that it is self-organizing as well as competitive. Generally, various models have been introduced for unsupervised learning tasks (e.g. ART1, ART2, etc.) or supervised ones (e.g. ARTMAP). The basic ART model is unsupervised and consists of various layers: 


{\bfseries $F_1$ Layer}: this layer denotes the input layer and propagates the input samples to the $F_2$ layer via bottom-up long-term memory units (LTMs) $\theta^{bu}$. As in the feedback mode, this layer provides the information being compared to the expectation of $F_2$'s output (via a top-down LTM $\theta^{td}$), it is also known as comparison layer.

{\bfseries $F_2$ Layer}: this layer yields the network output $y^{F_2}$ and serves as  competitive layer for categories. The LTM associated with a specific category $j$ is described with $\theta_j = \{\theta_j^{bu}, \theta_j^{td}\}$ 

{\bfseries Orienting Subsystem}: this module acts as control mechanism by inhibiting or allowing categories to resonate.

Generally, a sample $x$ is fed to the network and a winner-takes-all competition over the categories takes place at $F_2$ w.r.t to some objective function (e.g. similarity metrics). Afterwards, the orienting subsystem determines the adequacy of the selected category. According to a threshold (vigilance parameter $\rho$), either the system selects the best category and the system enters a resonance state and adapts the assigned LTM units. If the orienting subsystem rejects the category, a new one is created to encode the new sample. Hence, the vigilance parameter helps to incorporate new memories or new information. Higher vigilance produces more detailed memories, lower vigilance produces more general memories. For the interested reader, we would like to refer to the survey about various ART models presented by Silva et al. \cite{silva-2019-artSurvey}.



In the following, we will introduce our architecture which is designed similar to the ART models but where the weights are updated automatically within an AutoEncoder model. Because of the adaptive response of our new AutoEncoder model, we refer to it as the "Adapti\textbf{v}e AutoEn\textbf{coder}". 


\subsection{Terminology of KG}
The term 'knowledge Graph' has been emerged as a buzz word when speaking of applications on complex networks, and hence, the terminology blurs the line to other related topics terms like \emph{Knowledge Base} or \emph{Ontology}.

Ontologies themselves describe semantic modeling of knowledge and inherits not only the classes and properties but we can also have realizations (instances) of the ontology comprising entities. Due do the linkages defined by the ontology, the system gains semantic information. 
Even though this might already cover large parts where the term knowledge graph has been used in the literature, a characteristic of the KG is its reasoning capabilities which demarcates it from pure ontologies, resp., knowledge bases. The reasoning module is used to gain new (semantic) information.

In our work, we shine a light on the semantic linkage, therefore, we are interested in the underlying ontology of the graph at hand where our reasoning module is defined in chapter \ref{sec:adaptiveAutoEncoder}. Formally, we can describe the knowledge graph by a set of entities $\mathcal{E} = \{e_1, \ldots, e_{N_e}\}$ and a set of relation types $\mathcal{R} = \{r_1, \ldots, r_{N_r}\}$. A triple $(h, r, t)$ models the interaction between the entities $h:head$ and $t:tail$ by the relation type (predicate) $r$. 
\section{Adaptive AutoEncoder}
\label{sec:adaptiveAutoEncoder}

\newcolumntype{L}[1]{>{\hsize=#1\hsize\RaggedRight} X}
\begin{table}
\caption{Notation table}
\begin{tabularx}{\columnwidth}{L{0.1}|L{0.8}}
\toprule
\bfseries Symbol & \bfseries Description\\
\midrule
$h, t$ & head entity ($h$), tail entity ($t$) \\
$r$   & relation type \\
$\mathcal{N}(r)$ & adjacent nodes of relation $r$\\
$\mathcal{N}(x)$ & adjacent relations of entity $x$\\
$\mathbbm{1}_x$ & binary encoding for a nodes' x incident relations\\
$\Phi(\mathbbm{1}_x)$ & fingerprint of a node's incident relations\\
\bottomrule
\end{tabularx}
\end{table}

From cognitive science we know that inhibitory control allows us to control our attention, thoughts and leads to our behavior of how we act. It enables us also to suppress impulses from our environment which might tempt our disposition and lead to old habits and to conditioned responses. Therefore, inhibitory control allows us to adapt to our environment such that we can change our behavior, enables us to think about our habits and gives us last but not least choices. 
This phenomenon leads us to the terminology of \emph{Lateral Inhibition} which describes that a neuron's response to a stimulus is inhibited by the excitation of a neighboring neuron. In cognitive science, this is further studied in the so-called 'stop-signal' paradigm \cite{logan}. In general, an individual performs a 'go' task after an imperative stimulus occurred and a 'stop' signal leads to an inhibition of the individuals' response.



In this spirit we present an architecture suppressing signals to neighboring neuron within the same layer to create a bottleneck where information can pass through. In the next section, we discuss the \emph{V-Coder} more formally.

\subsection{V-Coder}
\subsubsection{\bfseries Logic of ARTs in V-Coder}\hfill\\
\label{sec:v_coder_art}
The V-Coder inherits the 2-fold idea from ART in a way such that the F1 layer (accepts and transforms the input data) is described in our model as the \emph{Encoder} module. The output of our encoder model is then processed in a way that resembles the functionality of the F2 layer. We pass the output to single neurons in the hidden layer between Encoder and Decoder of an AutoEncoder. Whereas in the vanilla AE this layer acts as the description of the input vectors in latent space, we interpret it in the V-Coder as the competitive layer, where only single neurons are activated. This inherits the idea of cognitive control mechanism. Likewise to traditional ART models, where the candidate neuron can now learn the input pattern, in our model, the information can also just flow through these selected neurons. In comparison to ART, where the following vigilance test would decide upon the clustering, we measure the reconstruction loss in order to decide whether the input can be assigned to the selected neuron on the competitive layer or not.
The idea is that for similar input data, the output of the encoder is similar, and therefore, a neuron in the competitive layer can be selected which subsume these similar input patterns. 
The information which is passed from the encoder module to the selected neuron on the competitive layer is used for the reconstruction in the \emph{Decoder} module. For similar input data, the reconstruction loss is fairly low compared to input data where we have a lot of variance from the encoding step.

\subsubsection{\bfseries Encoding of input data}\hfill\\
In the area of knowledge graphs our input data consists of facts. For the sake of brevity, we will discuss our approach on static knowledge bases where we have triples $(head, predicate, tail)$. Nevertheless, we would like to mention that if the semantic of a link is not changing over time, the approach could also be applied to non-static knowledge bases, also-called temporal knowledge graph (tKG), which can be represented by a set of quadruples (time-dependent facts) $G = \{s, r, o, t |s, o \in \mathcal{E}, r \in \mathcal{R}\, t \in \mathcal{T}\}$, where $\mathcal{E}$ is the set of entities, $\mathcal{R}$ is a set of relations and $\mathcal{T}$ denotes the temporal domain. In this case the predicate links two entities $subject$ and $object$ only at a certain point in time or within a time interval. 

As described above, the first step of our model is to compute a lower-dimensional representation, which we will refer to as a \emph{fingerprint} of an entity $x$. The role of an entity in a knowledge graph is described by its incident relations $\mathcal{N}(x) = \{r | \exists y \in \mathcal{E} \ldotp (x,r,y)\}$. For example, an unknown entity with incident relations \emph{'acts'},\emph{'nominee'}, \emph{'awarded\_to'} let us infer that the entity is most probably a person working in film industry as an actor. 

We can construct a binary vector representation $\mathbbm{1}_{x} \in \{0,1\}^{|\mathcal{R}|}$ for an entity by its incident relations, which we define as: 
\begin{equation}
    \mathbbm{1}_{x} = 
    \begin{cases} 1 &\text{if } r_k \in \mathcal{N}(x)\\
                0 &\text{otherwise}
    \end{cases}
\end{equation}

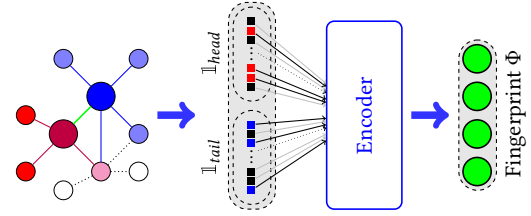
\begin{figure}
    \centering
     \scalebox{0.5}{\begin{tikzpicture}
	\tikzstyle{rec}=[minimum width=2cm, minimum height=1cm, rounded corners=5mm, dashed, thick, fill=gray!20, draw, anchor=west]
	\tikzstyle{neuron}=[circle, draw, very thick, minimum size=0.75cm,inner sep=0pt, anchor=center]
	\tikzstyle{mini_neuron}=[circle, draw, very thick, minimum size=0.25cm,inner sep=0pt, anchor=center]

    \node[draw, minimum size=0.75cm, circle, fill=purple] (n0) at (1, -0.5) {};
    \node[draw, minimum size=0.5cm, circle, fill=red] (n1) at (0, 0) {};
    \node[draw, minimum size=0.5cm, circle, fill=magenta!50] (n2) at (2, -1.5) {};
    \node[draw, minimum size=0.75cm, circle, fill=blue] (n3) at (2, 0.5) {};
    \node[draw, minimum size=0.5cm, circle, fill=blue!50] (n4) at (3, -0.5) {};
    \node[draw, minimum size=0.5cm, circle, fill=blue!50] (n5) at (1, 1.5) {};
    \node[draw, minimum size=0.5cm, circle, fill=white] (n6) at (3, -1.5) {};
    \node[draw, minimum size=0.5cm, circle, fill=red] (n7) at (0, -1.5) {};
    \node[draw, minimum size=0.5cm, circle, fill=white] (n8) at (1, -2) {};
    \node[draw, minimum size=0.5cm, circle, fill=blue!50] (n9) at (3, 1.5) {};

    \draw[-, thick, purple] (n0) -- (n1);
    \draw[-, thick, purple] (n0) -- (n7);
    \draw[-, thick, purple] (n0) -- (n2);
    
    \draw[-, very thick, green] (n0) -- (n3);
    
    \draw[-, thick, blue] (n3) -- (n2);
    \draw[-, thick, blue] (n3) -- (n4);
    \draw[-, thick, blue] (n3) -- (n5);
    \draw[-, thick, blue] (n3) -- (n9);
    
    \draw[dotted, thick] (n2) -- (n6);
    \draw[dotted, thick] (n2) -- (n8);
    \draw[dotted, thick] (n2) -- (n4);

    \draw[->, line width=2mm, blue!80] (3.5,0) -- (4.5,0);

    \draw node[rec, minimum width=1.25cm, minimum height=5.5cm] at (5.375,0.25) (indicators) {};
    
    \draw node[rec, minimum width=1cm, minimum height=2.5cm, label={[rotate=90, shift={(-1.25,0.75)}]above:\Huge $\mathbbm{1}_{head}$ }] at (5.5,1.625) (indicator_rels) {};
    \draw node[fill=black] (I-1) at (6, 0.75) {};
    \draw node[fill=red] (I-2) at (6, 1) {};
    \draw node[fill=red] (I-3) at (6, 1.25) {};
    \draw node[dotted] (I-4) at (6, 1.75) {\huge$\vdots$};
    \draw node[fill=black] (I-5) at (6, 2) {};
    \draw node[fill=red] (I-6) at (6, 2.25) {};
    \draw node[fill=black] (I-7) at (6, 2.5) {};
    
    \draw node[rec, minimum width=1cm, minimum height=2.5cm, label={[rotate=90, shift={(-1.25,0.75)}]above:\Huge $\mathbbm{1}_{tail}$ }] at (5.5,-1.125) (indicator_time) {};
    \draw node[fill=blue] (T-1) at (6, -0.25) {};
    \draw node[fill=black] (T-2) at (6, -0.5) {};
    \draw node[fill=blue] (T-3) at (6, -0.75) {};
    \draw node[dotted] (T-4) at (6, -1.) {\huge$\vdots$};
    \draw node[fill=black] (T-5) at (6, -1.5) {};
    \draw node[fill=black] (T-6) at (6, -1.75) {};
    \draw node[fill=blue] (T-7) at (6, -2.0) {};

    \node[draw, minimum width=5cm, minimum height=2cm, text=blue, rounded corners=2mm, draw=blue, very thick, rotate=90] at (9,0) (Ext_fn) {\Huge Encoder};
    
    \draw[->, black!25] (I-1) -- (Ext_fn);
    \draw[->, thick] (I-2) -- (Ext_fn);
    \draw[->, thick] (I-3) -- (Ext_fn);
    \draw[->, black!25] (I-5) -- (Ext_fn);
    \draw[->, thick] (I-6) -- (Ext_fn);
    \draw[->, black!25] (I-7) -- (Ext_fn);
    \draw[->, dotted] (I-4) -- (Ext_fn);

    \draw[->, thick] (T-1) -- (Ext_fn);
    \draw[->, black!25] (T-2) -- (Ext_fn);
    \draw[->, thick] (T-3) -- (Ext_fn);
    \draw[->, black!25] (T-5) -- (Ext_fn);
    \draw[->, black!25] (T-6) -- (Ext_fn);
    \draw[->, thick] (T-7) -- (Ext_fn);
    \draw[->, dotted] (T-4) -- (Ext_fn);

    \draw node[rec,minimum width=1cm, minimum height=4cm, label={[rotate=90, shift={(-1.75,-0.5)}]right:\Huge Fingerprint $\Phi$}] at (11.5,0) (fingerprint) {};
    \draw node[neuron, fill=green] at (12, -1.5) {};
    \draw node[neuron, fill=green] at (12, -0.5) {};
    \draw node[neuron, fill=green] at (12, 0.5) {};
    \draw node[neuron, fill=green] at (12, 1.5) {};

    \draw[->, line width=2mm, blue!80] (10.25,0) -- (11.25,0);

    
\end{tikzpicture}}  

    \caption{Fingerprint of a triple (h,r,t)}
    \label{fig:fingerprint}
\end{figure}

After the relational information of the head and tail entity of a fact (h,r,t) is translated to binary vectors, we concatenate them $\mathbbm{1}_{h}||\mathbbm{1}_{t}$ and use them as input for a deep forward network, an encoder parameterized by $\theta_{enc}$, yielding the fingerprint for the topology, denoted by $\Phi_{h,t}$. In other words, the function $\Phi: U \rightarrow R^d$ describes a non-linear transformation from the space $U \subseteq \mathbbm{1}_{h} \times \mathbbm{1}_{t}$ to a lower $d$-dimensional vector in the real vector space. Hence, the highly sparse representation are transformed to a dense lower dimensional space. 
The procedure is illustrated in figure \ref{fig:fingerprint}.

\subsubsection{\bfseries Lateral Inhibition}\hfill\\
\label{sec:lateral_inhibition}
A vanilla AutoEncoder is designed such that we impose a bottleneck in the network forcing to compress our original input data into a lower-dimensional space. If the input data inherits some sort of structure, e.g., correlation between input feature, the model is able to learn it. However, if the input features are independent from each other, the compression task and therefore the reconstruction in the decoder is a difficult task. 

In our architecture, we can think of the layer between the encoder and decoder - which in the AE case computes the latent representation of the input data - as the competitive layer as they have been used in ART architectures. Hence, the activation of the competitive layer is defined by a sigmoid function to squash the receiving data to the range [0,1]. The neuron with the maximal value is the one being considered as the only active one, therefore, transforming the idea of lateral inhibition from the field of cognitive science into our architecture. All other neurons are considered as inactive. 

More formally the competitive layer with lateral inhibition receives the input data from the previous layer. Recall, that the previous layer is the output of the encoder module, therefore, we are receiving the fingerprint $\Phi$. For each neuron in the competitive layer we are calculating the activations in a straightforward way:
\begin{equation}
    h_j = \sigma(\sum w_{ij}\Phi_i),
\end{equation}
 where the weights $w_{ij}$ denote the weights connecting the encoder's output with the competitive layer and $\sigma(\cdot)$ defining an activation function. In the same spirit as the F2-layer of ART networks, we define the winner neuron $j$ in an unsupervised learning setting as the one with the highest score: $argmax_{0\leq j \leq k} \sigma(\sum w_{ij}\Phi_i)$. 
 
The winner neuron is now the only neuron the information can pass through, therefore, is responsible for the reconstruction of the received data. All other neurons are inhibited. 

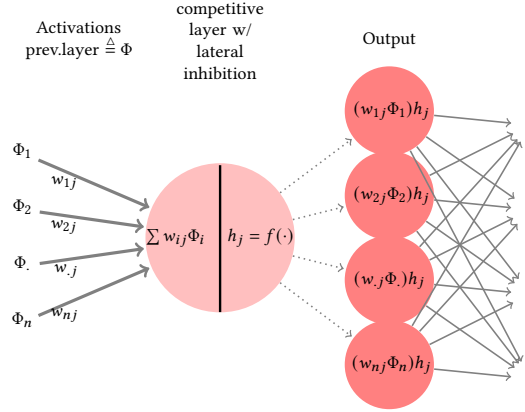
\begin{figure}
    \centering
     \scalebox{0.75}{\def\layersep{2.0cm}

\begin{tikzpicture}[shorten >=1pt,->,draw=black!50, node distance=\layersep]
    \tikzstyle{every pin edge}=[<-,shorten <=1pt];
    \tikzstyle{every pin thick_edge}=[<-,shorten <=1pt, ultra thick];
    \tikzstyle{neuron}=[circle,fill=red!25,minimum size=17pt,inner sep=0pt];
    \tikzstyle{mini_neuron}=[circle,fill=black!25,minimum size=45pt,inner sep=0pt];
    \tikzstyle{input neuron}=[neuron, fill=white!50];
    \tikzstyle{output neuron}=[neuron, fill=blue!50];
    \tikzstyle{hidden mu}=[neuron, fill=orange!50];
    \tikzstyle{hidden sigma}=[neuron, fill=black!50];
    \tikzstyle{hidden neuron}=[neuron, fill=gray!50];
    \tikzstyle{activated neuron}=[neuron, fill=red!50];
    \tikzstyle{hidden mini_neuron}=[mini_neuron, fill=red!50];
    \tikzstyle{annot} = [text width=6em, text centered];
    
    \tikzset{
        cross/.style={path picture={ 
        \draw[black, very thick, -] (path picture bounding box.south) -- (path picture bounding box.north);
    }}
    } 

    \foreach \name / \y / \annot in {1/1/1,2/2/2,3/3/.,4/4/n}
        \path[yshift=2.5cm]
            node[input neuron, label={[yshift=-0.5cm]$\Phi_\annot$}] (I-\name) at (0,-\y) {};

    \node[neuron, cross] (unit) at (1.75*\layersep, 0) {$\sum w_{ij} \Phi_i $ \quad $ h_{j} = f(\cdot)$};

    \foreach \name / \y in {1/0.75,2/2.25,3/3.75,4/5.25}
        \path[yshift=3.cm]
            node[hidden mini_neuron] (M-\name) at (3.25*\layersep,-\y) {
            \ifthenelse{\name = 3}{$(w_{. j}\Phi_.) h_j$}{
                \ifthenelse{\name = 4}{$(w_{n j}\Phi_n) h_j$}
                    {$(w_{\name j}\Phi_\name) h_j$}}};
            
    \foreach \name / \y in {1/1.5,2/3,3/4.5,4/6}
        \path[yshift=3.5cm]
            node[input neuron] (O-\name) at (4.5*\layersep,-\y) {};


    \foreach \src in {1,...,4}
        \path (I-\src) edge[ultra thick] node[near start, yshift=-0.2cm] {
            \ifthenelse{\src = 3}
                {$w_{. j}$}
                {\ifthenelse{\src = 4}
                    {$w_{n j}$}
                    {$w_{\src j}$}
                }
            }
            (unit);
    
    \foreach \dest in {1,...,4}
        \path (unit) edge[thick, dotted] (M-\dest);
        
    \foreach \src in {1,...,4}
        \foreach \dest in {1,...,4}
            \path (M-\src) edge[thick] (O-\dest);

    \node[annot,above of=unit, node distance=3.5cm] (max_select) {competitive layer w/ lateral inhibition};
    \node[annot,left of=max_select, xshift=-0.5cm] {Activations prev.layer $\overset{\triangle}{=} \Phi$};
    \node[annot,right of=max_select, xshift=1cm] {Output};

\end{tikzpicture}}  

    \caption{Node of the V-Coder}
    \label{fig:node_vCoder}
\end{figure}

In order to reconstruct the input data, we now calculate the \emph{Hadamard} product of the received fingerprint with the weights connecting it to the winner neuron. The score of the winner neuron is used as scaling parameter. Therefore, we define the input to the decoder module - parameterized by $\theta_{dec}$ - as (cf. figure \ref{fig:node_vCoder}):
\begin{equation}
\label{eq:input_decoder}
    v_j = (W_{:,j} \odot\Phi)h_j,
\end{equation}

Finally, The output layer yields the reconstruction $z_j=Dec(v_j)$ where the mean squared error (MSE) defines the reconstruction loss. As the information passed solely one node in the competitive layer, we define it as a conditioned loss:
\begin{equation}
J ( x, z | j) = \|x - z_j\|^2,
\end{equation}
where $j$ denotes the active unit in the competitive layer and $z_j$ defines the reconstruction based upon the information it received according to equation \ref{eq:input_decoder}.


\begin{figure}
    \centering
     \scalebox{0.75}{\def\layersep{2.0cm}

\begin{tikzpicture}[shorten >=1pt,->,draw=black!50, node distance=\layersep]
    \tikzstyle{every pin edge}=[<-,shorten <=1pt]
    \tikzstyle{every pin thick_edge}=[<-,shorten <=1pt, ultra thick]
    \tikzstyle{neuron}=[circle,fill=black!25,minimum size=17pt,inner sep=0pt]
    \tikzstyle{mini_neuron}=[circle,fill=black!25,minimum size=6pt,inner sep=0pt]
    \tikzstyle{input neuron}=[neuron, fill=green!50];
    \tikzstyle{output neuron}=[neuron, fill=blue!50];
    \tikzstyle{hidden mu}=[neuron, fill=orange!50];
    \tikzstyle{hidden sigma}=[neuron, fill=black!50];
    \tikzstyle{hidden neuron}=[neuron, fill=gray!50];
    \tikzstyle{activated neuron}=[neuron, fill=red!50];
    \tikzstyle{hidden mini_neuron}=[mini_neuron, fill=red!50];
    \tikzstyle{annot} = [text width=4em, text centered];
    \tikzstyle{rect} = [rectangle, fill=orange!25];

    \foreach \name / \y in {1,...,5}
        \node[input neuron, pin=left:] (I-\name) at (0,-\y) {};

   	\draw node[minimum width=1cm, minimum height=4cm, rounded corners=1mm, dashed, very thick, draw, anchor=west, fill=gray!10,] at (\layersep-0.5cm,-3) {};
    \foreach \name / \y in {1,...,4}
        \path[yshift=-5.5cm]
            node[hidden mu] (H_enc-\name) at (\layersep, \y cm) {};

    \foreach \name / \y in {1,...,3}
        \path[yshift=-5cm]
            node[hidden neuron] (H-\name) at (\layersep+\layersep, \y cm) {};
    
    \path[yshift=-5cm] node[activated neuron] (H-2) at (\layersep+\layersep, 2 cm) {};
            
    \foreach \name / \y in {1/.25, 2/.5, 3/.75, 4/1}
        \path[yshift=-5cm]
            node[hidden mini_neuron] (M-\name) at (\layersep+\layersep+0.5cm, 1.375cm + \y cm) {};

   	\draw node[minimum width=1cm, minimum height=4cm, rounded corners=1mm, dashed, very thick, draw, anchor=west, fill=gray!10,] at (3*\layersep-0.5cm,-3) {};
    \foreach \name / \y in {1,...,4}
        \path[yshift=-5.5cm]
            node[hidden mu] (H_dec-\name) at (\layersep+\layersep + \layersep,\y) {};

    \foreach \name / \y in {1,...,5}
            \node[output neuron, pin={[pin edge={->}]right:}] (O-\name) at (4*\layersep,-\y) {};

    \foreach \source in {1,...,5}
        \foreach \dest in {1,...,4}
            \path (I-\source) edge (H_enc-\dest);

    \foreach \source in {1,...,4}
        \foreach \dest in {1,...,3}
            \path (H_enc-\source) edge (H-\dest);

    \foreach \source in {1,...,4}
        \path (H_enc-\source) edge[ultra thick] (H-2);

    \foreach \source in {1,...,4}
        \foreach \dest in {1,...,4}
        \path (M-\source) edge (H_dec-\dest);

    \foreach \source in {1,...,4}
        \foreach \dest in {1,...,5}
            \path (H_dec-\source) edge (O-\dest);


    \node[annot,above of=H-1, node distance=3.5cm] (max_select) {Competitive Layer};
    \node[annot,left of=max_select] (encoder) {Encoder};
    \node[annot,left of=encoder] (input) {Input};
    \node[annot,right of=max_select] (decoder) {Decoder};
    \node[annot,right of=decoder] (output) {Output};

\end{tikzpicture}}  

    \caption{Architecture of the V-Coder}
    \label{fig:architecture_vCoder}
\end{figure}
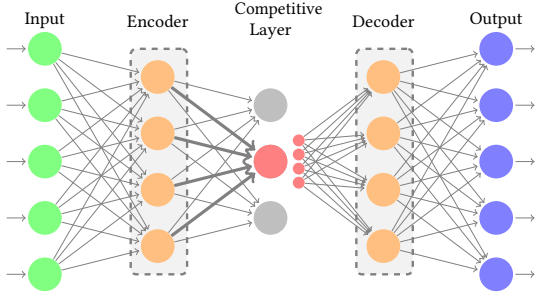

\subsubsection{\bfseries Adaptive Step}\hfill\\
In our setting of semantic disclosure in a knowledge graph, we integrate a supervision upon the competitive layer. We exploit the fact that there is a limited expressiveness when the information flows through single nodes in the competitive layer. If the input data, i.e., the fingerprints, do not yield any correlation within their latent features, then we conclude that the topologies of the original entities are too diverse. In other words, similar entities have common incident relation types, and therefore, the encoder yields similar fingerprints for those entities.
The supervision we integrate in the V-Coder is the relation type connecting two entities. Hence, the number of neurons in the competitive layer are the number of relation types $|\mathcal{R}|$ in the KG.
Considering the graph in figure \ref{fig:fingerprint}, the edge connecting the two entities refers to a specific unit in the competitive layer. Using the additional supervision in the application of semantic disclosure, we get a quality measure for a fingerprint w.r.t a specific relation by passing the information through the respective unit in the competitive layer. Consequently, we get the information which relation has to capture too many information in the sense that it has to express too diverse fingerprints. A high variance in the reconstruction is therefore an indicator that a relation is too general and needs a further inspection in the underlying ontology. 
Likewise to traditional ART networks, we extend the size of the competitive layer such that the information results in different neurons (cluster), i.e., we extend the semantic of a relation. 
Let $C=\{c_0, \ldots, c_j, \ldots, c_m\}$ be the units in the competitive layer.
Suppose that for the unit $c_j$ we receive the highest variance w.r.t the reconstruction loss. The V-Coder selects this unit and extends the layer by one additional unit $c_{m+1}$. Then it receives again data whose supervision was formerly conditioned on $c_{j}$, we now allow the information to select the best one amongst the set $\{c_j, c_{m+1}\}$ in the manner described in \ref{sec:lateral_inhibition}, i.e., we select as winner neuron the one resulting in the minimal reconstruction loss:

\begin{equation}
\argmin_{c_j \in C} J ( x, Dec(v_j) | c_j) = \|x - Dec(v_j)\|^2,
\end{equation}

The adaption step, i.e., increasing the number of neurons in the competitive layer (\emph{plasticity}) can be done without touching any other information learned so far within the network. Note that the encoder module ($\theta_{enc}$) and decoder module ($\theta_{dec}$) are preserved, as well as the weights from the output of the encoder to the neurons in the competitive layer (\emph{stability}). We extend the weight matrix connecting the encoder with the competitive layer by one additional dimension $W_{:,m+1}=W_{:,j}$ which creates the linkage to the new unit in the competitive layer (c.f. clustering ART models).

\section{Evaluation}
\label{sec:evaluation}

\subsection{Settings}
\subsubsection{\bfseries Datasets}\hfill\\
For our evaluation, we use three benchmark datasets:

\noindent

\begin{itemize}
    \item \textbf{FB15k-237} \cite{toutanova15conv} was created from $\textit{FB15k}$ by removing the inverse of many relations from the training set as well as from the validation and test set which makes the link prediction task more difficult
    \item \textbf{YAGO3-10} \cite{Mahdisoltani-2015-yago} is a subset of YAGO3 consisting of entities that have a minimum of $10$ relations each. It has $123.182$ entities and $37$ relations. Most of the triples deal with descriptive attributes of people, such as 'citizenship', 'gender', 'profession' or 'marital status'
    \item \textbf{NELL-995} \cite{xiong-etal-2017-deeppath} is a subset of the 995-th iteration of NELL
\end{itemize}


\noindent The statistics of the datasets are summarized in table \ref{tab:datasets}.

\begin{table}[t]
    \centering
    \resizebox{1\columnwidth}{!}{
    \begin{tabular}{l|r|r|r|r|r}
         Dataset & \# $N_e$ & \# $N_r$ & \#train & \#test & \#valid \\
         \hline \hline
         FB15k-237 & 14.541 & 237 & 272.115 & 20.466 & 17.535\\
         YAGO3-10 & 123.182 & 37 & 1.079.040 & 5.000 & 5.000\\
         NELL-995 & 75,492 & 200 & 149.678 & 3992 & 543 \\
         \bottomrule
    \end{tabular}
    }
    \caption{Dataset statistics}
    \label{tab:datasets}
\end{table}

\subsubsection{\bfseries Implementation Details}\hfill\\
All experiments were implemented in Python 3.7.3 with PyTorch 1.2.0. For the computation of the link prediction we use the \emph{OpenKE} framework \cite{han2018openke} which is an open-source package for the computation of knowledge graph embeddings.

\subsubsection{\bfseries Experimental Setup}\hfill\\
In our evaluation we use a single hidden layer for the encoder module of dimension \{$16,32$\} encoding the fingerprint of the input triple. The reconstruction in the decoder module uses also one hidden layer equally sized. The number of neurons in the competitive layer are given by the number of relational types within a dataset. 

We run the clustering procedure for $20$ epochs. After that, the cluster neuron with the highest variance in the reconstruction loss is selected and splitted into two units whilst preserving the learned information of other nodes. 
For tackling bias in the new cluster nodes, we use an exponential decay function \begin{equation}
    r = (\epsilon_{start} + (\epsilon_{end} - \epsilon_{start})) * exp(-t\cdot \epsilon_{decay}),
\end{equation}
where $\epsilon_{start}$, $\epsilon_{end}$ and $e_{decay}$ (decay factor) are hyperparameters and $t$ is the current iteration number. After a splitting operation, this heuristic initially explores both units to be the best one for the input data and introduces some randomness till it homes in to the one with minimal reconstruction loss. This procedure is beneficial due to the unknown cluster size, resp., semantic rewirings in the knowledge graph. 
We set the learning rate of the V-Coder from $\{0.01,0.001,0.0001\}$ with a batch size in the range of $\{16, 32, 64\}$, a weight decay of $0.001$ and the epsilon values of: $\epsilon_{start}=1.0$, $\epsilon_{end}=0.01$, $\epsilon_{decay} \in \{1e-4, 1e-5, 1e-6\}$. As optimizer we are using ADAM \cite{kingma14adam}.

\subsection{Analytical Evaluation}
\subsubsection{\bfseries Effectiveness of V-Coder}\hfill\\
The first part of the evaluation are hand-crafted experiments to show evidence that the proposed V-Coder indeed separates links according to their inherently semantic structure within the knowledge graph. 
For that, we randomly select links within the datasets FB15K-237 and NELL-995 to create a copy where 2 links are merged. Hence, we loss relational types by mixing up the semantic information. Just according to the incident relations towards entities being linked to the new created links, the task for the V-Coder is to uncover again the original relations and separate them such that they reproduce the linkages of the original dataset. 
In the competitive layer of the V-Coder we pass therefore additional information through the selected neuron being now in charge for both original semantics. As the merging operation might not result in the highest corruption within the dataset (which might be also of low sample rates within the input), we tell the V-Coder which neuron in the competitive layer to split. After the splitting operation, the V-Coder learns again the assignments of the input data. 
The results for various recovery tasks are shown in table \ref{tab:recovery}. For example, when merging the two relations with ids $124$ (/education/educational\_institution/school\_type) and $16$ (/organization/organization/headquarters./location/mailing\_address/country) from FB15K-237 in a new dataset, the V-Coder reproduces the linkages for the id $124$ with $96.29\%$ accuracy and for $16$ with $85.39\%$, resulting in an overall reproducibility of $90.84\%$. Note that the quality of the recovery is dependent on the sample size in the original dataset. The V-Coder recognized different fingerprints which are then passed to different neurons in the competitive layer such that we can split the hand-made relation again into the original ones. 

\newcommand*{\MinNumber}{0.0}%
\newcommand*{\MidNumber}{50.0} %
\newcommand*{\MaxNumber}{100.0}%

\newcommand{\ApplyGradient}[1]{%
        \IfDecimal{#1}{
        \ifdim #1 pt > \MidNumber pt
            \pgfmathsetmacro{\PercentColor}{max(min(100.0*(#1 - \MidNumber)/(\MaxNumber-\MidNumber),100.0),0.00)} %
            \hspace{-0.33em}\colorbox{green!\PercentColor!yellow}{#1}
        \else
            \pgfmathsetmacro{\PercentColor}{max(min(100.0*(\MidNumber - #1)/(\MidNumber-\MinNumber),100.0),0.00)} %
            \hspace{-0.33em}\colorbox{red!\PercentColor!yellow}{#1}
        \fi}
        {#1}
}
\newcolumntype{R}{>{\collectcell\ApplyGradient}c<{\endcollectcell}}

\begin{table}
    \centering
    \resizebox{1\columnwidth}{!}{
    \begin{tabular}{r|| *{4}{R}|  |*{4}{R}}
        & \multicolumn{4}{c}{Recovery - FB15K-237} & \multicolumn{4}{c}{Recovery NELL-995}\\  
        \toprule
        \midrule
         & (124,16) & (80, 11) & (56,226) & Avg. & (110,181) & (70,152)& (31,119) & Avg.\\
        \midrule
        $rel_1$ & 96.3 & 75.5 & 93.4 & 88.4 &     71.4 & 54.2 & 72.8 & 66.1\\
        $rel_2$ & 85.4 & 92.6 & 79.1 & 85.7 &     56.0 & 66.3 & 71.7 & 64.6\\
        \midrule
        $Avg$   & 90.8 & 84.0 & 86.2 & 87.0 &     63.7 & 60.2 & 72.3 & 65.3\\
        \bottomrule
    \end{tabular}    
    }
    \caption{Examples of accuracy of recovered links after artificially merged (c.f. appendix \ref{app:semantics})}
    \label{tab:recovery}
\end{table}

\subsubsection{\bfseries Error Reduction}\hfill\\
\begin{figure*}[!ht]
    \begin{minipage}{.35\textwidth}
    \centering
    \includegraphics[width=\linewidth]{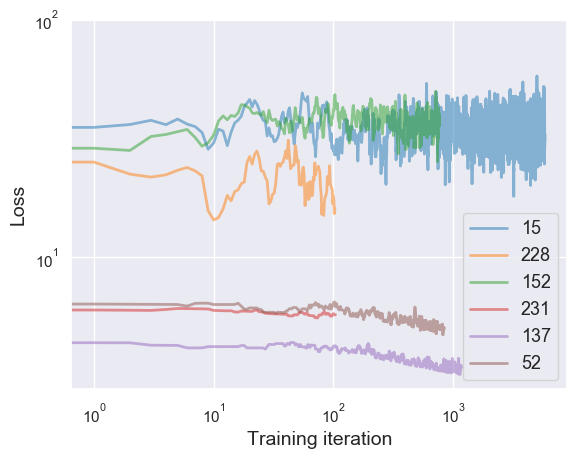}
    \caption{Reconstruction loss on individual relations [15: 'location/contains', 228: '/split\_to', 152: '/combatants', 231: '.../seasonal\_months', 137: '.../nutrient' , 52: '/music/.../parent\_genre']}
    \label{fig:loss_examples_fb15k237}    
    \end{minipage}
    \begin{minipage}{.64\textwidth}
    \centering
        \begin{minipage}{.4\textwidth}
            \centering
            \includegraphics[width=\linewidth]{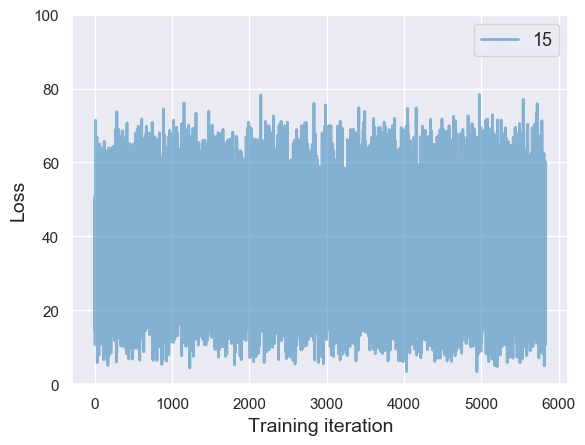}
        \end{minipage}
        \begin{minipage}{.4\textwidth}
            \centering
            \includegraphics[width=\linewidth]{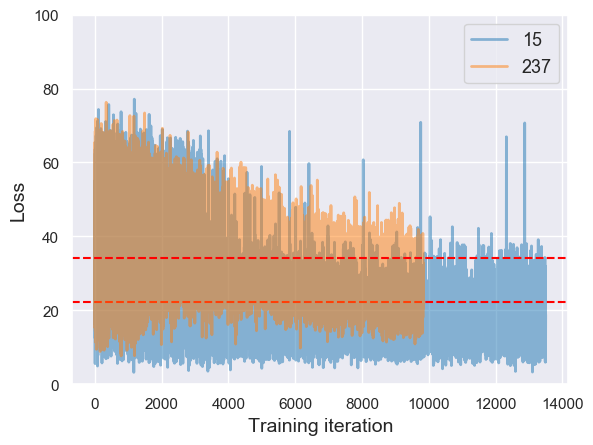}
        \end{minipage}
        \begin{minipage}{.4\textwidth}
            \centering
            \includegraphics[width=\linewidth]{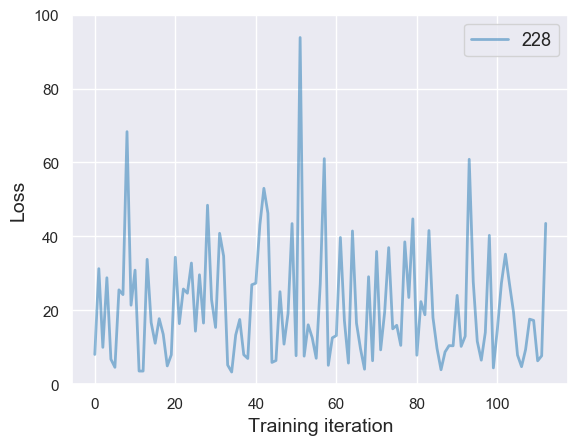}
        \end{minipage}
        \begin{minipage}{.4\textwidth}
            \centering
            \includegraphics[width=\linewidth]{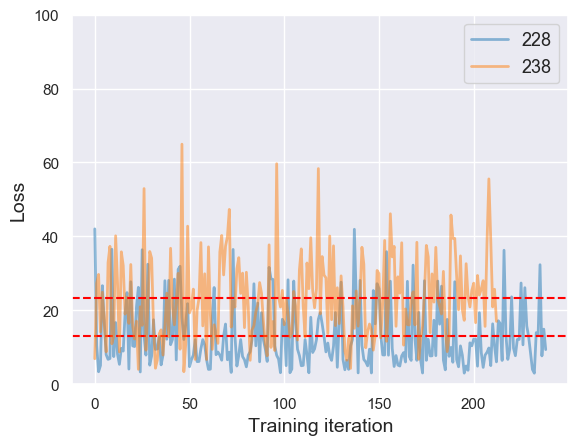}
        \end{minipage}
    \caption{Reconstruction loss after splitting operation in competitive layer; Link 15 splitted in [15, 237] (top), Link 228 splitted in [228, 238] (bottom). Red lines indicate averages of the losses.}
    \label{fig:loss_post_split_fb15k237}
    \end{minipage}
\end{figure*}

\begin{figure*}
    \begin{minipage}{.35\textwidth}
    \centering
    \includegraphics[width=\linewidth]{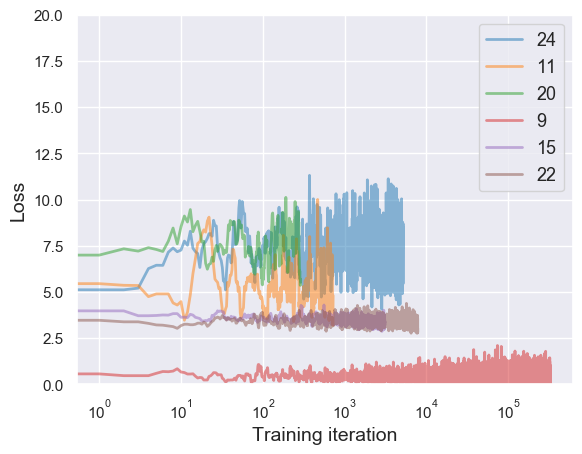}
    \caption{Reconstruction loss on individual relations [24: participatedIn , 11: owns, 20: hasOfficialLanguage, 9: playsFor, 15: edited, 22: hasMusicalRole]}
    \label{fig:loss_examples_yago}    
    \end{minipage}
    \begin{minipage}{.64\textwidth}
    \centering
        \begin{minipage}{.4\textwidth}
            \centering
            \includegraphics[width=\linewidth]{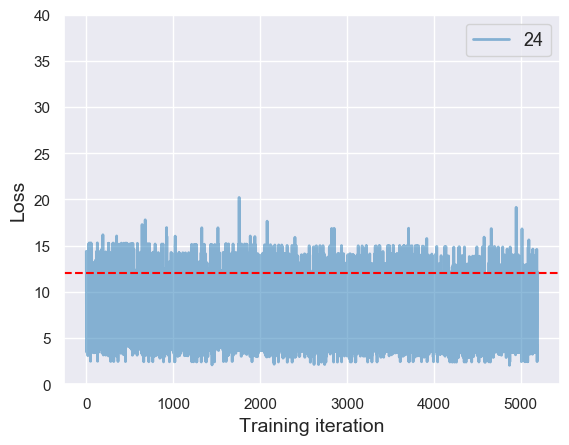}
            \label{fig:loss_examples}    
        \end{minipage}
        \begin{minipage}{.4\textwidth}
            \centering
            \includegraphics[width=\linewidth]{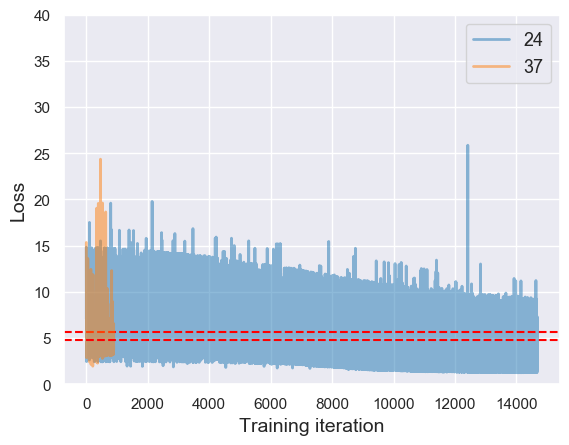}
            \label{fig:loss_examples}    
        \end{minipage}
    \caption{Reconstruction loss after splitting operation in competitive layer; Link 24 splitted in [24, 37]. Red lines indicate averages of the losses. }
    \label{fig:loss_post_split_yago}    
    \end{minipage}
\end{figure*}

Next, we discuss the individual error reduction achieved by the V-Coder. For our discussion we use the FB15K-237 and the YAGO3-10 datasets. Analysing the individual error terms, we can observe that some fingerprints, and therefore, characteristics of the knowledge graph can be learnt effortless. 
A low reconstruction loss results from the fact that similar input topologies have similar fingerprints
. Apart from that we also observe links where the variance of the reconstruction loss is high which results from topologies where the fingerprints are too diverse to be expressible by one active neuron. This is illustrated in the case of FB15K-237 in figure \ref{fig:loss_examples_fb15k237} and for YAGO3-10 in figure \ref{fig:loss_examples_yago}. The reason for that lies in the underlying dataset, more precisely, in the ambiguous semantic a link carries. 
In the illustration we see that the relation with the id $15$ is highly difficult to reconstruct which is the relation "/location/location/contains". The ambiguity results from the fact that "contains" is a generalization interlinking entities from various domains. For example, "contain" might describe that a city is "contained" within a state, a university is 'contained' within a city, or a federal state is "contained" within a country. Therefore, this relation can be regarded as a candidate for a semantic disclosure and the question is if the disentangling helps in reducing the reduction error. The answer is illustrated in \ref{fig:loss_post_split_fb15k237}. 
In our example, the link with id "15" is chosen to be splitted as it exceeds its capabilities to encompass all the input data which is directed to it. 
On top of fig \ref{fig:loss_post_split_fb15k237}, we see that after splitted into the relations with the ids 15,237, the mean loss is reduced leading to a better description of the relations. Also in the case of the relation with id $228$ being splitted into the relation $228, 238$, we observe a denoising of the reconstruction loss, hence, reducing the variance in the reconstruction.

\subsubsection{\bfseries Link Prediction Results}\hfill\\
In this section, we show the effects on the link prediction task.

Our evaluation metrics follows the standard on link prediction tasks, i.e, we use mean reciprocal rank (MRR) and \emph{Hits@k}, with $k \in \{1, 10\}$. Mean reciprocal rank denotes the average of the inverse of the mean rank assigned to the true triple over all candidate triples, whereas \emph{Hits@k} measures the percentage of times a true triple is ranked within the top-$k$ candidate triples.

We choose state-of-the-art embedding methods being also suitable for large knowledge graphs like YAGO3-10. For the single embeddings methods we choose the learning rate from \{0.0001, 0.001, 0.01, 0.1\}, the weight decay from \{0.001, 0.001, 0.01, 0.1\} and $d$ for the dimensionality of the latent space from \{100, 150, 200\}. We train the models using \emph{Adam} \cite{kingma14adam} and \emph{Adagrad} \cite{duchi2011adaptive} with a batch size ranging in $\{64, 128, 256\}$.

The results are summarized in table \ref{tab:link_prediction_fb_nell_yago}, where we show the results on the original datasets as well as on the datasets after passing the V-Coder, i.e., after a splitting of the relations has been executed. Note that, for the run on the extended datasets, we fix all hyperparameters having been used on the original dataset. The suffixes in the datasets denote the number of splits having been made.
Note that we are not feeding the system with an increased number of samples at hand to boost the link prediction dramatically., but we arrange it new w.r.t the new semantics. Nevertheless, as we can see, through the separation of relations we can observe small improvements on the various datasets. These improvements can be explained by the re-arranged features in latent space as well as the additional expressiveness of the gained relations. Hence, disentangling the manifolds in the low-dimensional space helps for the link prediction itself.

\begin{table*}[t]
    \centering
    \resizebox{\textwidth}{!}{
    \begin{tabular}{l| cc|cc|cc c cc|cc|cc c cc|cc|cc}
         & \multicolumn{6}{c}{FB15k-237 \& FB15K-237-10splits} && \multicolumn{6}{c}{NELL-995 \& NELL-995-10splits} && \multicolumn{6}{c}{YAGO3-10 \& YAGO3-10-1split} \\ 
         \cmidrule{1-7} \cmidrule{9-14} \cmidrule{16-21}
         & \multicolumn{2}{c}{MRR} & \multicolumn{2}{c}{Hits@10} & \multicolumn{2}{c}{Hits@1} &&
         \multicolumn{2}{c}{MRR} & \multicolumn{2}{c}{Hits@10} & \multicolumn{2}{c}{Hits@1} && 
         \multicolumn{2}{c}{MRR} & \multicolumn{2}{c}{Hits@10} & \multicolumn{2}{c}{Hits@1} \\
         \cmidrule{1-7} \cmidrule{9-14} \cmidrule{16-21}
         DistMult \cite{yang2015distmult} & .250 & \eq{.250} & .392 & \up{.394} & .178 & \eq{.178}  && 
         .221 & \up{.247} & .321 & \up{.352} & .160 & \up{.179} && 
         .316 & \up{.317} & .522 & \up{.523} & .212 & \up{.213} \\
         TransE\cite{bordes13transe} & .287 & \up{.288} & .476 & \up{.477} & .190 & \up{.191} &&
         .301 & \up{.302} & .429 & \up{.431} & .222 & \eq{.222} && 
         .264 & \up{.265} & .469 & \eq{.469} & .158 & \up{.160} \\
         RotatE\cite{sun18rotate} & 
         .251 & \eq{.251} & .426 & \up{.427} & .164 & \down{.163} && 
         .334 & \up{.338} & .414 & \up{.423} & .288 & \down{.287} && 
         .143 & \down{.142} & .252 & \up{.254} & .083 & \up{.084} \\
         Complex\cite{trouillon16complex} & .260 & \up{.264} & .435 & \up{.442} & .174 & \eq{.174} &&
         .267 & \up{.285} & .356 & \up{.383} & .216 & \up{.227} && 
         .409 & \down{.407} & .605 & \up{.609} & .310 & \down{.306} \\
         Analogy\cite{liu2017analogy} & .301 & \up{.302} & .467 & \up{.468} & .219 & \eq{.219} && 
         .142 & \up{.150} & .266 & \up{.277} & .091 & \up{.092} && 
         .345 & \eq{.345} & .550 & \up{.553} & .241 & \up{.241} \\
         SimplE\cite{kazemi-2019-simplE} 
         & .283 & \up{.285} & .445 & \up{.447} & .202 & \up{.204} &&
         .223 & \eq{.223} & .318 & \up{.323} & .173 & \eq{.173} && 
         .481 & \up{.498} & .656 & \up{.663} & .389 & \up{.411} \\
    \bottomrule
    \end{tabular}
    }
    \caption{Effects on link prediction result on \textit{FB15K-237}, \textit{NELL-955} and \textit{YAGO3-10}.}
    \label{tab:link_prediction_fb_nell_yago}
\end{table*}

\subsection{Semantic Evaluation}
In figure \ref{fig:loss_post_split_fb15k237}, we show that the V-Coder identifies the relation with id $15$:'location/location/contains' in the dataset FB15K-237 as the one with the highest variance on the reconstruction loss. 
Not only our common-sense already signals that 'contains' can connect various entities from different domains, but also the V-Coder identifies this relation as being ambiguous in the sense that the fingerprints created by the encoder vary such that leading the information through one neuron on the competitive layer we lose too much information for the reconstruction. Hence, to tackle the ambiguity, the V-Coder creates a new unit which creates a new cluster in the relational domain. In table \ref{tab:semantic_disclosure} we show in decreasing order the number of entities which are now captured by two units on the competitive layer. We see that the original relation $15$ indeed related entities from different domains such as countries, states, rivers, cities but also entities for colleges/universities (not shown in the table). Hence, the V-Coder enriches the knowledge graph automatically-driven by a new semantic relation splitting the entities up such that similar fingerprints are interlinked by the relations with id $15$ and a new relation with id $237$ (notice that the ids are zero-based). What we can observe is that the V-Coder now tunnels countries well-separated through the unit $237$, whereas states and counties are captured by the unit $15$. In conclusion, the V-Coder clusters similar fingerprints on country-level on one side, and similar fingerprints resulting from state-level on the other side and enriches therefore the knowledge graph by a new semantic relation. This follows our intuition of the original relation, where a sub-level in the ontology could be extended by the relations 'contains\_state', 'contains\_city', 'contains\_county', etc. 

\begin{table}
    \centering
    \resizebox{1\columnwidth}{!}{
    \begin{tabular}{r|r|r|r}
         \multicolumn{4}{c}{15:location/location/contains}
         \\
         \midrule \midrule
         \multicolumn{2}{c}{Relation: 15} & \multicolumn{2}{c}{Relation: 237}\\
         \midrule
         \multicolumn{1}{c}{heads} & \multicolumn{1}{c}{tails} & \multicolumn{1}{c}{heads} & \multicolumn{1}{c}{tails} \\
         \midrule
         ('California',145) & ('Washington County',9) & ('USA',948) & ('Richmond', 6)\\
         ('New York', 98) & ('Richmond', 8) & ('UK', 267) & ('Hamilton', 6) \\ 
         ('Eurasia', 93) & ('Concord', 7) & ('England', 229) & ('Springfield', 4) \\ 
         ('Europe', 77) & ('Lake County', 5) & ('Canada', 89) & ('Halifax', 4) \\ 
         ('Massachusetts', 66) & ('Jefferson County', ) & ('Germany', 72) & ('Rhine', 4) \\ 
         \vdots & \vdots & \vdots & \vdots \\ 
         \bottomrule
    \end{tabular}
    }
    \caption{Semantic Disclosure in FB15K-237 of relation 15:location/location/contains}
    \label{tab:semantic_disclosure}
\end{table}

\section{Conclusion}
\label{sec:conclusion}

In this work, we introduced a novel adaptive model, called the V-Coder. It inherits the unsupervised learning idea from \emph{Adaptive Learning Theory} (ART) and wraps it up in an architecture similar to an AutoEncoder. 
We showed that the V-Coder is able to identify relations which inherently connects entities from various domains. Those relations can be considered as being ambiguous and are candidates for disentanglement. For that, a competitive layer carries the idea of lateral inhibition which suppresses information to flow through neighboring nodes. Upon the variance in the reconstruction loss, the information is inferred which (cluster) neuron is overloaded with information, and hence, yields an indicator to split this (cluster) neuron into two units. Therefore, the V-Coder adaptively changes their size on the competitive layer while learning the clusters information. In our application on semantic disclosure in knowledge graphs, we show a first application of the V-Coder. The evaluation shows that the V-Coder is able to recover semantic links from corrupted input data. By enriching a knowledge graph with new semantic information, we can show the tendency to improve link prediction tasks on the benchmark datasets FB15K-237, YAGO3-10 and NELL-995. This augmentation is justified by semantic reasoning based on the existence of independent semantic clusters for a relation being present in the input knowledge graph. 
For future works, we would like to extend the idea of V-Coder also to other architectures and applications for dynamic neural nets.



\bibliographystyle{ACM-Reference-Format}
\balance
\bibliography{refs}

\appendix
\section{Relational descriptions}
\label{app:semantics}
The semantics of the relations being used throughout the paper are as follows:
\begin{table*}[t]
    \centering
    \resizebox{0.9\textwidth}{!}{
    \begin{tabular}{l||r|l}
         Dataset & id & Description\\
         \hline \hline
         \multirow{12}{*}{FB15k-237} & 11 & /film/film/release\_date\_s./film/film\_regional\_release\_date/film\_release\_distribution\_medium\\
         & 15 & /location/location/contains\\
         & 16 & /organization/organization/headquarters./location/mailing\_address/country\\
         & 52 & /music/genre/parent\_genre\\
         & 56 & /location/hud\_county\_place/place\\ 
         & 80 & /sports/sports\_team/sport\\
         & 124 & /education/educational\_institution/school\_type\\
         & 137 & /food/food/nutrients./food/nutrition\_fact/nutrient\\
         & 152 & /military/military\_combatant/military\_conflicts./military/military\_combatant\_group/combatants\\
         & 226 & /organization/organization\_founder/organizations\_founded\\
         & 228 & /dataworld/gardening\_hint/split\_to\\
         & 231 & /base/localfood/seasonal\_month/produce\_available./base/localfood/produce\_availability/seasonal\_months\\
         \midrule
         \multirow{6}{*}{NELL-995} & 
         31 & concept:topmemberoforganization\\
         & 70 & concept:buildinglocatedincity\\
         & 110 & concept:automakerproducesmodel\\
         & 119 & concept:headquarteredin\\
         & 152 & concept:countryalsoknownas\\
         & 181 & concept:radiostationincity\\
         \midrule
         \multirow{5}{*}{YAGO3-10} & 9 & playsFor \\
         & 11 & owns \\
         & 15 & edited \\
         & 22 & hasMusicalRole \\
         & 24 & participatedIn
    \end{tabular}
    }
    \caption{Semantic of relations}
    \label{tab:datasets}
\end{table*}

\end{document}